\documentclass[conference]{IEEEtran}
\IEEEoverridecommandlockouts
\usepackage[latin1]{inputenc}
\usepackage{amsmath}
\usepackage{amsfonts}
\usepackage[normalem]{ulem}
\usepackage{color}
\usepackage{tabularx}
\usepackage{booktabs}
\usepackage{float}
\usepackage{amssymb}
\usepackage{graphicx}
\usepackage[export]{adjustbox}[2011/08/13]
\usepackage{enumerate}
\usepackage{caption}
\usepackage{array}
\usepackage{multirow}
\usepackage{tabulary}
\usepackage{colortbl}
\usepackage[bottom]{footmisc}
\usepackage[latin1]{inputenc}
\usepackage{dsfont}
\newcommand\independent{\protect\mathpalette{\protect\independenT}{\perp}}
\def\independenT#1#2{\mathrel{\rlap{$#1#2$}\mkern2mu{#1#2}}}
\usepackage[english]{babel}
\renewcommand{\thesubsection}{\thesection.\alph{subsection}}
\usepackage{graphics}
\usepackage{hyperref}
\usepackage{scalerel,stackengine}
\allowdisplaybreaks
\stackMath
\newcommand\reallywidehat[1]{%
	\savestack{\tmpbox}{\stretchto{%
			\scaleto{%
				\scalerel*[\widthof{\ensuremath{#1}}]{\kern-.6pt\bigwedge\kern-.6pt}%
				{\rule[-\textheight/2]{1ex}{\textheight}}
			}{\textheight}%
		}{0.5ex}}%
	\stackon[1pt]{#1}{\tmpbox}%
}
\makeatletter
\newcommand*{\indep}{%
	\mathbin{%
		\mathpalette{\@indep}{}%
	}%
}
\newcommand*{\nindep}{%
	\mathbin{
		\mathpalette{\@indep}{\not}
	}%
}
\newcommand*{\@indep}[2]{%
	\sbox0{$#1\perp\m@th$}
	\sbox2{$#1=$}
	\sbox4{$#1\vcenter{}$}
	\rlap{\copy0}
	\dimen@=\dimexpr\ht2-\ht4-.2pt\relax
	\kern\dimen@
	{#2}%
	\kern\dimen@
	\copy0 
} 
\makeatother
\begin{document}

\title{Estimating Heterogeneous Causal Effects in the Presence of Irregular Assignment Mechanisms\\
}

\author{\IEEEauthorblockN{Falco J. Bargagli Stoffi}
\IEEEauthorblockA{\textit{IMT School for Advanced Studies, Lucca, Italy}\\
falco.bargaglistoffi@imtlucca.it}
\and
\IEEEauthorblockN{Giorgio Gnecco}
\IEEEauthorblockA{\textit{IMT School for Advanced Studies, Lucca, Italy}\\
giorgio.gnecco@imtlucca.it}
}
\maketitle
\thispagestyle{plain}
\pagestyle{plain}

    \begin{abstract}

	This paper provides a link between causal inference and machine learning techniques - specifically, Classification and Regression Trees (CART) - in observational studies where the receipt of the treatment is not randomized, but the assignment to the treatment can be assumed to be randomized (irregular assignment mechanism). 
	The paper contributes to the growing applied machine learning literature on causal inference, by proposing a modified version of the Causal Tree (CT) algorithm 
	to draw causal inference from an irregular assignment mechanism. The proposed method is developed by merging the CT approach with the instrumental variable framework to causal inference, 
	hence the name Causal Tree with Instrumental Variable (CT-IV). As compared to CT, the main strength of CT-IV is that it can deal more efficiently with the heterogeneity of causal effects, as demonstrated by a series of numerical results obtained on synthetic data. 
	Then, the proposed algorithm is used to evaluate a public policy implemented by the Tuscan Regional Administration (Italy), which aimed at easing the access to credit for small firms. 
	In this context, CT-IV breaks fresh ground for target-based policies, identifying interesting heterogeneous causal effects.

\end{abstract}

\vspace{-0.4cm}

\section{Introduction} \label{Intro}

   Modern statistics is experiencing the growth of machine learning techniques, such as Classification and Regression Trees (CART)  [\ref{breiman1984}], and Random Forests [\ref{breiman2001}],
    which can be applied to a wide range of statistical problems. In order to use these techniques
    to answer relevant statistical questions, it is appropriate to highlight some important features
    of many machine learning methods.
    These methods are largely about
    making good predictions and finding the model that fits the data best. Furthermore, their importance lies in the ability to deal with complex datasets, where the number of units is large, as well as the number of features connected with a single unit. 
     In this framework, causality is often de-emphasized. However, in the last decades, the availability of increasingly larger datasets has brought to the attention a new important problem for causal inference, which machine learning techniques can ``easily'' solve.
    As a matter of fact, the necessity to deal with problems connected with the heterogeneity of the treatment effects is stronger than in the past:
		the availability of large datasets makes it possible to customize causal effect estimates for population's subsets and even for individuals.
    In the past, the analysis subsets for causal inference problems were specified in advance by trials' protocols, while with the new machine learning technique presented in this paper, the subsets are selected by the algorithm itself in a data-driven way.
    Classical approaches to the analysis of heterogeneous effects are non-parametric methods,
    such as nearest neighbour matching method, kernel method, and series estimation [\ref{wagerathey2017}].
    These techniques usually offer good results in terms of estimation abilities.
    The drawback is that they perform well as far as the number of covariates is low.
    Machine learning techniques outperform other non-parametric methods when the number of covariates is relatively high.
    This can be seen as the reason that led recently to the application of machine learning techniques to causal discovery and inference. 
    	A good example of the use of machine learning techniques in these fields, and a very important inspiration for the present work, is the 
			recently published paper 
			[\ref{atheyimbens2015}], 
			where an adaptation to causal inference of CART in its regression version, named Causal Tree (henceforth, CT), was developed to estimate causal effects with instrumental variables. While the goal of the method proposed in [\ref{atheyimbens2015}] is very similar to the one of the algorithm we develop in this paper (namely, to draw proper causal inference in the presence of irregular assignment mechanisms), 
			the CT algorithm can identify the heterogeneity of causal effects with respect to a particular subset of selected covariates, where the selection needs to be done by the researcher herself. Conversely, our algorithm, named Causal Tree with Instrumental Variable (henceforth, CT-IV), provides a data-driven way to shed light on the heterogeneity of the treatment effects. 
		 The paper is structured as follows. Section \ref{Causal} provides a background on the causal inference framework, 
		its link with machine learning 
		as it is modeled via the CT algorithm, and basic concepts about instrumental variables. In Section \ref{unconfounded}, we describe our proposed CT-IV algorithm. In Section \ref{synth}, we provide a comparison on synthetic data between the CT and CT-IV algorithms in the presence of an irregular assignment mechanism, showing numerically advantages of the latter. Section \ref{assess} concludes the paper with a case study on firm level data where the proposed algorithm is used to assess the heterogeneity of the effects of an employment policy implemented by the Tuscan Regional Administration (Italy). 
		\vspace{-0.25cm}
	
	\section{Background} \label{Causal}
    \textbf{A. Rubin's Causal Model.} In order to set up the method presented in this paper, it is important to remind some notions and notations of Rubin's potential outcome framework [\ref{rubin1974}, \ref{rubin1980}]. 
    Rubin's framework is the milestone of causal inference. Together with the Pearl's causality approach [\ref{pearl2009}], it is the most widely used model in the scientific literature about causal inference.
    
    Given a set of $N$ units, indexed by $i=1,...,N$, let $W_i$ be the binary indicator of the receipt of the treatment:
    \begin{equation}\label{eq:1} \small
    W_i \in \{0,1\}.
    \end{equation}
    In order to develop a proper causal inference framework, one needs to assume that the potential outcomes for any unit do not vary with the treatments assigned to other units, and that, for each unit,
    there are no different forms or versions of each treatment level, which may lead to different potential outcomes [\ref{rubin1974}, \ref{rubin1980}]. This assumption is referred in the literature as the Stable Unit Treatment Value Assumption (henceforth, SUTVA).
    Given SUTVA, one can postulate the existence of a pair of potential outcomes for each unit:
    \begin{equation} \small
    Y_i^{obs}= Y_i(W_i)=\begin{cases} Y_i(0) & \text{if} \;\;\;\;  W_i=0, \\ Y_i(1) &  \text{if} \;\;\;\;  W_i=1.
    \end{cases}
    \end{equation}
    Starting from the notion of potential outcomes, one can define a unit-level causal effect as the difference between the potential outcome
    under treatment and the one under control:
    \begin{equation} \small
    \tau_i=  Y_i(1) -  Y_i(0).
    \end{equation}
    The problem of this approach to causal inference is that one can observe
    just one potential outcome for every unit. 
    It is impossible to observe both potential outcomes for the same unit at the same time.
    Therefore, from this perspective, causal inference is a \textit{missing data problem} [\ref{ir2015}]. 
    
	Is it then impossible to estimate any causal effect? No, it is not but, in orded to draw proper causal inference, one needs to introduce the central concepts of the Rubin's Causal Model [\ref{ir2015}].
    		Let $X_i$ be the vector of features (usually called also \textit{covariates} or \textit{pre-treatment} variables) associated with the $i$-th unit, and known not to be affected by the treatment. Let $\bold{X}$ be the $N \times K$ matrix of covariates values (where $N$ is the number of units, and $K$ the number of covariates per unit), ${W}$ the $N$-dimensional vector of binary assignments to the treatment, and ${Y\text{(0)}}$ and ${Y\text{(1)}}$ the $N$-dimensional vectors of potential outcomes.
    Imbens and Rubin [\ref{ir2015}] define the \textit{assignment mechanism} $P({W}|\bold{X}, {Y\text{(0)}}, {Y\text{(1)}})$, 
			the \textit{unit level assignment probability} $p_i(\bold{X},{Y\text{(0)}}, {Y\text{(1)}})$ 
		and the \textit{propensity score} $e(x)=P(W_i=1|X_i=x)$, which is the probability for a unit to be treated, conditional on its covariates [\ref{rosenbaum1983}].
    
    Following 
		[\ref{ir2015}], one defines a \textit{classical randomized experiment} as an assignment mechanism that has the following 4 properties:
\begin{enumerate}
	\item it is \textit{individualistic}, meaning that the treatment assignment for any unit is a function only of its own covariates and potential outcomes; 
	\item it is \textit{probabilistic}, meaning that the unit level assignment probability belongs to the open interval $(0,1)$;
	\item it is \textit{unconfounded}, meaning that it does not depend on the potential outcomes;
	\item it has a functional form that is \textit{known} (and, to some extent, controlled) by the researcher.
\end{enumerate}
Suppose that one is interested in the population average treatment effect:
\begin{equation} \small
\tau^p=\mathbb{E}[Y_i(1)-Y_i(0)]={  \mu}(1) - {  \mu}(0),
\end{equation}
where ${  \mu}(1)$ is the expected value of $Y_i(1)$, and ${  \mu}(0)$ is the expected value of $Y_i(0)$.
In the case of a classical randomized experiment, an unbiased estimator of $\tau^p$ is:
\begin{equation} \small
	\hat{\tau} = \hat{  \mu}(1) - \hat{  \mu}(0).
\end{equation}
In the equation above, $\hat{  \mu}(1) = {1 \over N_1} \sum_{i:W_i=1}Y_i^{obs}$, where $N_1$ is the number of units assigned to the treated group, and $\hat{  \mu}(0) = {1 \over N_0} \sum_{i:W_i=0}Y_i^{obs}$, where $N_0$ is the number of units assigned to the control group. Finally, $\sum_{i\in \{0,1\}}N_i=N$.

By relaxing the fourth property of a known assignment mechanism, one ends up in a scenario that 
[\ref{ir2015}] defines as a \textit{Regular Assignment Mechanism}. Is it possible in such a scenario to still draw causal inference? The central property that needs to be invoked in order to do so is the unconfoundedness property 3) defined above. Unconfoundedness can be formalized as the conditional independence of the assignment variable $W_i$
to the potential outcomes given (conditioning on) the covariates vector: 
\begin{equation}\label{eq:unconf} \small
W_i \independent \left( Y_i(0), Y_i(1)\right) | X_i.
\end{equation}
The importance of this assumption is that, conditional on covariates, one can treat observations as they were coming from a randomized experiment. Let the Conditional Average Treatment Effect (CATE) be defined as:
\begin{equation} \small
\tau(x)=\mathbb{E}[Y_i(1)-Y_i(0)|X_i=x] = \mu(1, x) - \mu(0, x), \label{cate}
\end{equation}
where $\mu(w, x)$ is the expected value of $Y_i(W_i=w)$ given $X_i=x$. Then it can be proven, by the law of iterated expectations, that:
\begin{equation} \small
\tau^{p}=\mathbb{E}[Y_i(1)-Y_i(0)]=\mathbb{E}[\tau(X_i)]=\mathbb{E}[\mu(1, x) - \mu(0, x)].
\end{equation}
It follows that $\tau^p$ is identified 
if $\mu(1,x)$ and $\mu(0,x)$ are identified over the support of $\bold{X}$. Under unconfoundedness, it can be proven that $\mu(1,x)$ and $\mu(0,x)$ are identified [\ref{atheyimbens2015}]. 
This gives the possibility to the researcher, if all the important confounding covariates 
are present in the data, to draw causal inference even when the assignment mechanism is not randomized but is regular. This is the typical case of observational studies, where the researcher does not know beforehand the assignment mechanism (i.e., property 4) above does not hold). 
Moreover, in observational studies, 
the assignment to the treatment may be different from the receipt of the treatment. In this scenario, where one allows for non-compliance between the treatment assigned and the treatment received,  one can assume that the assignment is itself unconfounded, while the receipt is confounded. 
Following 
[\ref{ir2015}], this assignment mechanism is named \textit{Irregular Assignment Mechanism}. How to draw inference in the presence of an irregular assignment mechanism will be the focus of Subsection II.C, and also the focus of our applied machine learning algorithm in Section \ref{unconfounded}.

Going back to the CATE, there is a variety of reasons for researchers to conduct estimation of $\tau(x)$ (see formula (\ref{cate})).
One is strictly related to the magnitude of the benefits of the treatment which can vary with the features of the individuals.
For instance, one can imagine the extreme case where the average treatment effect of a drug is positive on the overall population
(in terms of curing a specific disease), but for a sub-population of patients, with certain characteristics, the average treatment effect is ineffective, or even negative.
For these reasons, it is important to find a proper way to estimate causal effects not only on the entire population, but also on specific subsets of the population. 

\vspace{0.3cm}
		
    \textbf{B. Regression Trees for Causal Inference.} 
    Machine learning offers new ways to investigate
    heterogeneous effects (i.e., ones that depend on the covariates vector $X_i$, see (\ref{cate})), as suggested in 
		[
		\ref{atheyimbens2015},
		\ref{wagerathey2017}]. Machine learning techniques developed so far in the literature can provide a useful tool to achieve this goal, in scenarios where the assignment mechanism is randomized or is regular.
		
		A machine learning technique that was applied to this task is the CART method [\ref{breiman1984}]. CART is suitable for this goal because, on one side, it is a fully supervised machine learning technique but, on the other side, it is a pretty flexible method that can be adapted to various learning tasks. Here, due to page constraints, we limit to provide an overview of the basic ideas behind such method, referring the reader to [\ref{breiman1984}] for other details about it. 
		The primary goal of CART is to estimate the conditional expectation of an observed outcome
		on the basis of the information on features and outcomes for units in the training sample, and to compare the resulting estimates on a test sample. Practically, one can estimate these values by building a 
		suitable tree (a classification or a regression tree, depending on the specific problem). 
		The different admissible 
		tree models one can construct entail alternative splits of the tree, based on the values of the features in the data. A possible way to choose the best among various admissible 
		trees is provided by the following procedure, whose initial step consists in dividing the dataset into two different samples: \\
    	\textit{a)} a first sample, called training sample (or training set), which is used to construct a maximal depth tree, performing the splits using an in-sample goodness-of-fit measure $Q^{is}$. The size of this training sample is indicated by $N^{tr}$. Then, the maximal depth tree is pruned, with the aim of maximizing another criterion function $Q^{crit}$, for various choices of a suitable penalty parameter $\alpha > 0$ on which $Q^{crit}$ depends;\\
    	\textit{b)} a second sample, called validation sample (or validation set), which is used, for each choice of $\alpha$, to validate the associated pruned tree, through the use of an out-of-sample goodness of fit $Q^{oos}$. This second sample size is indicated by $N^{va}$. 
    	
    	Here, we consider the case in which a single training set and a single validation set are used. In the machine learning literature, this procedure is called the holdout method, and is particular form of cross-validation. 
		In this case, $\alpha$ is chosen by maximizing $Q^{oos}$ with respect to it, and the tree itself is re-trained using the full dataset, for the resulting value of $\alpha$.  Finally, a different 
		sample, called test sample (or test set), with cardinality $N^{te}$, is used to assess the performance of the resulting model.

	    In the following, we describe the Causal Tree (CT) method [\ref{atheyimbens2015}], which is a modification of the original CART method in its regression version, tailored to causal inference.
		The CT method differs from CART from the following features:
    \begin{enumerate}[a.]
    	\item the CATE transformation of the outcome;
    	\item a rework of the in-sample goodness of fit;
    	\item a rework of the out-of-sample goodness of fit.
    \end{enumerate}
    
    \subsubsection{The CATE Transformation}
    First of all,
    one needs to address the big issue of constructing an algorithm
    that leads to an accurate estimate $\hat{\tau}(x)$ of the conditional average treatment effect.
    In an ideal world, one would measure the quality of the estimator by looking at the value of the following goodness of fit measure, defined in terms of the mean squared error:
    \begin{equation}\label{infeasible} \small
    Q^{infeas}(\hat{\tau})= - \mathbb{E}[(\tau_i-\hat{\tau}(X_i))^2],
   \end{equation}
    However, it is infeasible to estimate the value of $Q^{infeas}$, because one does not know the values of both potential outcomes for each unit, as $\tau_i$ is unobservable.
    To address this issue, one can transform the observed outcome using the treatment indicator variable $W_i$ and the propensity score $e(X_i)$, as proposed by Athey and Imbens [\ref{atheyimbens2015}]:
    \begin{equation}\label{eq:10} \small
    Y_i^*=Y_i^{obs} \cdot {{W_i- e(X_i)}\over{(1-e(X_i)) \cdot e(X_i)}}.
    \end{equation}
    Since $Y_i^{obs}$ is equivalent to $Y_i(W_i)$ then, using (\ref{eq:1}), one can express (\ref{eq:10}) as:
    \begin{equation} \label{eq:13} \small
    	Y_i^*=Y_i(1) \cdot {W_i \over e(X_i)} - Y_i(0) \cdot {{(1-W_i)} \over {(1-e(X_i))}}.
    \end{equation}
    What is the strength of this transformation? Athey and Imbens prove that, if
		the unconfoundedness assumption holds, then:
    \begin{equation}\label{eq:12} \small
    \mathbb{E}[Y_i^*|X_i=x]= \tau(x),
    \end{equation}
    where $Y_i^*$ in (\ref{eq:13}) is computed replacing the propensity score $e(X_i)$ with its suitable estimate $\hat{e}(X_i)$ (obtained, e.g., via logistic regression). 
		However, there are some issues in building a tree using a straightforward transformation of the outcome like $Y_i^*$. In fact, Athey and Imbens argue that the within a leaf sample average of the transformed outcome $Y_i^*$ is not the most efficient estimator of the treatment effect and, moreover, that the proportion of treated and control units within a leaf can be quite different from the overall sample proportion.
    An easy way to solve this issue, proposed in 
		[\ref{atheyimbens2015}], is to weight the CATE transformation in a matter similar to the one developed in 
		[\ref{hirano2003}]. Every partition of the covariates space is identified by a set of leaves, and
    the treatment effect for the covariates vector $X_i$ belonging to a generic leaf $\mathbb{X}_j$ is estimated as\footnote{Likewise next formulas (\ref{IV}) and (\ref{IV_propensity}), (\ref{cace_leaf}) can be applied also to the validation sample and to the entire (training and validation) sample $\Omega$, replacing the superscript ``\textit{tr}'', respectively, with ``\textit{va}'' and ``$\Omega$''.}:
    \begin{eqnarray} \label{cace_leaf} \small
    \hat{\tau}^{CT}(X_i)&=&{{\sum_{l:X_l^{tr} \in \mathbb{X}_j}  Y_l^{obs,tr} \cdot {W_l^{tr} \over \hat{e}(X_l^{tr})}} \over {\sum_{l:X_l^{tr} \in \mathbb{X}_j}{W_l^{tr} \over \hat{e}(X_l^{tr})}}} \nonumber \\
		&&- {{\sum_{l:X_l^{tr} \in \mathbb{X}_j}  Y_l^{obs,tr} \cdot {(1-W_l^{tr}) \over (1-\hat{e}(X_l^{tr}))}} \over {\sum_{l:X_l^{tr} \in \mathbb{X}_j}{(1-W_l^{tr}) \over (1-\hat{e}(X_l^{tr}))}}}.
    \end{eqnarray}

    \subsubsection{In-Sample Goodness of Fit}
    The second component of the algorithm, which also differs from the corresponding component in the original CART algorithm,
    is the in-sample goodness of fit.
    The big issue for defining a proper criterion function for the in-sample goodness of fit is that, in the causal inference framework, the criterion (\ref{infeasible}), and even its sample approximation 
		$-{1\over N^{tr}} \sum_{i=1}^{N^{tr}}(\tau_i-\hat{\tau}(X_i^{tr}))^2$,
	which is what would be implemented by a direct application of the original CART algorithm, are infeasible.
	Hence, 
	[\ref{atheyimbens2015}] proposes to approximate (\ref{infeasible}) by:
    \begin{equation} \small
    Q^{is}=-{1\over N^{tr}}\sum_{i=1}^{N^{tr}} \hat{\tau}^2(X_i^{tr}),
    \end{equation}
		and to use the corresponding criterion function
		\begin{equation} \small
    Q^{crit}=Q^{is}-\alpha \cdot \kappa,
    \end{equation}
		where $\alpha>0$ is a penalty parameter, and $\kappa$ is the number of leaves in the tree.  
    \subsubsection{Out-of-Sample Goodness of Fit}
    For cross-validation, there is no big need
    for any significant additional computational effort, given the fact that one has already obtained an estimate $\hat{\tau}^{CT}$ of the causal effect defined in terms of the training sample (see (\ref{cace_leaf})), 
		and one just needs to compare it with the causal effect drawn from the validation sample used for the cross-validation.
    One could easily rework the mean squared error with the transformed outcome $Y_i^*$ to get the Transform-The-Outcome (TOT) loss function:
    \begin{equation} \label{catebased} \small
    Q^{oos,tot}=-{1\over N^{va}} \sum_{i=1}^{N^{va}}( Y_i^{va,*}-\hat{\tau}(X_i^{va}))^2.
    \end{equation}


The in-sample goodness of fit can be reworked in different ways, following 
[\ref{atheyimbens2015}].	
It looks to us that the TOT-based out-of-sample goodness-of-fit in (\ref{catebased}) fits in a better way in those frameworks in which the number of covariates would lead to very computationally demanding alternative estimators. 
 
    \subsubsection{Causal Inference with Causal Tree}
		Due to the specific construction of the Causal Tree, the learning problem reduces to that of estimating the treatment effect in each member of a partition of the covariate space. For the problem of estimating the treatment effect in each leaf of the partition, standard methods are valid.
    Once one has constructed the tree $\mathbb{T}$, one can consider the leaf that corresponds
    to the subset $\mathbb{X}_j$ (henceforth, identified with $\mathbb{X}_j$ itself). The tree is defined as a partition of the feature space $\mathbb{X}$, and one can write:
    \begin{equation} \small
    \mathbb{T} = \{\mathbb{X}_1,..., \mathbb{X}_{\#(\mathbb{T})}\}, \: {\rm with} \: \bigcup_{j=1}^{\#(\mathbb{T})} \mathbb{X}_j = \mathbb{X},
    \end{equation}
    where $\#(\mathbb{T})$ indicates the number of leaves in the tree.
     Within the leaf $\mathbb{X}_j$, the average treatment effect is:
    \begin{equation} \small
    \tau_{\mathbb{T},\mathbb{X}_j}= \mathbb{E}[Y_i(1) - Y_i(0)|X_i \in
    \mathbb{X}_j],
    \end{equation}
    which  can be estimated as follows, by subtracting the average outcome $\overline{Y}_{\mathbb{X}_j}^{obs,te}(0)=\overline{Y}_c^{obs,te}$ on the control units from the average outcome $\overline{Y}_{\mathbb{X}_j}^{obs,te}(1)=\overline{Y}_t^{obs,te}$ on the treated units, both evaluated over the test sample,  which is different from the training and validation sample used for the cross-validation:
    \begin{equation} \small
    	\hat{\tau}_{\mathbb{T},\mathbb{X}_j}= \overline{Y}_{\mathbb{X}_j}^{obs,te}(1) - \overline{Y}_{\mathbb{X}_j}^{obs, te}(0)=\overline{Y}_t^{obs,te} - \overline{Y}_c^{obs, te}.
    \end{equation}
    One can also estimate, for each leaf $\mathbb{X}_j$, the variance of this estimator using the following Neyman estimator [\ref{neyman1934}]: 
     \begin{equation}\label{Neyman} \small
    \hat{\mathbb{V}}^{Neyman}_{\mathbb{T},\mathbb{X}_j}= { s^{te,2}_{t,{\mathbb{X}_j}} \over N^{te}_{t,{\mathbb{X}_j}}} + {s^{te,2}_{c,{\mathbb{X}_j}} \over N^{te}_{c,{\mathbb{X}_j}}},
    \end{equation}
    where $s^{te,2}_{t,{\mathbb{X}_j}}$ represents the sample variance of the treated group in the test set, $N^{te}_{t,{\mathbb{X}_j}}$
    its size, $s^{te,2}_{c,{\mathbb{X}_j}}$ the sample variance of the control group in the test set, and $N^{te}_{c,{\mathbb{X}_j}}$ its size.
    This estimator of the variance is unbiased, with respect to the finite-sample 
		distribution of the test sample, if the treatment effect can be assumed to be \textit{additive and constant} within a leaf [\ref{ir2015}].
    However, it can be used to construct confidence intervals only under the normal approximation, which is typically reliable when the number of units
    inside a leaf is large enough.

\vspace{0.3cm}
\textbf{C. General Instrumental Variable Framework.} \label{geniv}
    In observational studies, the assignment mechanism may be
    irregular. For example, dependence on the assignment of the
    potential outcomes may be not ruled out even after conditioning
    on a rich set of covariates. These are the cases where the unconfoundedness assumption is violated.
   In these settings, instrumental variable methods [\ref{ir2015}] can still help to
    estimate causal effects.
    To briefly make the context clear, one can consider the following example of an irregular assignment
    mechanism, for which, in a study population of $N$ units, a certain number of individuals are randomized to receive a treatment (read a drug), but not all the units that are assigned to receive it are actually treated.
    
Let us denote by $W_i$ the receipt of the treatment, 
and by $Z_i$ the assignment to the treatment (instrumental variable). Throughout this paper, we will assume both the $W_i$ and $Z_i$ to be binary, even if one could get similar results by relaxing this assumption. In the following, $W_i(Z_i)$ represents the treatment received as a function of the treatment assigned. This leads one to distinguish four different sub-populations $G_i$ of units: those that always comply with the assignment (compliers), those who never comply with the assignment (defiers), those that even if not assigned to the treatment take it (always-takers), and those who do not take the treatment even if assigned to it (never-takers). 
Formally, one can highlight these sub-populations as follows:
        \begin{enumerate} \small
    	\item Compliers ($G_i=C$): $W_{i}(Z_i=0)=0$ and $W_{i}(Z_i=1)=1$;
    	\item Defiers ($G_i=D$): $W_{i}(Z_i=0)=1$ and $W_{i}(Z_i=1)=0$;
    	\item Always-takers ($G_i=AT$): $W_{i}(Z_i=0)=1$, $W_{i}(Z_i=1)=1$;
    	\item Never-takers ($G_i=NT$): $W_{i}(Z_i=0)=0$, $W_{i}(Z_i=1)=0$.
    \end{enumerate} \normalsize

    \par How can one conduct causal inference in such a setting, if one decides to use the CART method? 
    The first thing to do is to assume the classical Instrumental Variable (IV) assumptions to hold [\ref{ir2015}]: monotonicity, existence of compliers, unconfoundedness of the instrument, and exclusion restriction.
    These four assumptions can be written in detail as follows:
    \begin{enumerate} 
    	\item monotonicity: $W_i (1) \geq W_i (0)$;
    	\item existence of compliers: $	P(W_{i}(0)<W_{i}(1))>0$;
    	\item unconfoundedness of the instrument (expressed in terms of conditional independence notation):
    	$Z_i \independent (Y_i(0,0), Y_i(0,1), Y_i(1,0),Y_i(1,1 ),W_i(0), W_i(1))$;
    	\item exclusion restriction: 
    	$Y_i(0) = Y_i(1),  \ \ {\rm for } \ G_i \in \{AT,NT\}$ where, for each sub-population and $z \in \{0,1\}$, the shortened notation $Y_i(z)$ is used to denote $Y_i(z, W_i(z))$.
    \end{enumerate}

    The monotonicity assumption leads us to exclude the existence of units that do exactly the opposite of what they are assigned to (read defiers). In the case of one-sided noncompliance, when units that are not assigned to take the drug cannot take it, this assumption is automatically satisfied as $W_i(0) = 0$ for each unit, excluding the presence of defiers and always-takers. In the case of two-sided noncompliance, when treated and control units can access the opposite treatment status, the monotonicity assumption is very plausible but not directly verifiable. The second assumption is the so-called ``existence of compliers'' assumption. This assumption states that the sub-population of compliers exists with positive probability. The third assumption states that the instrument is unconfounded. As we saw in Section \ref{Causal}, the importance of unconfoundedness is that, conditional on covariates, the assignment to the treatment is as good as if the assignment mechanism was randomized. The last but not least assumption is the exclusion restriction, which rules out any direct effect of $Z_i$ on $Y_i$. According to this assumpton, there is no effect of the assignment on the outcome, in the absence of an effect of the assignment of the treatment on the treatment received, being the treatment of primary interest.
    
    \indent 1) {\em Complier Average Causal Effect:} \label{CACE} In the setting above, what ``one can get from the data'' (without invoking any of the previous assumptions) is the Intention To Treat $ITT_Y$, which is defined as the effect of the intention to treat a unit on the outcome of the same unit (effect of the assignment):
    \begin{equation}\label{super_IV} \small
    	ITT_Y=\mathbb{E}[Y_i|Z_i=1] - \mathbb{E}[Y_i|Z_i=0].
    \end{equation}
			
If one does not assume any of the classical IV assumptions above 
to hold, then the global $ITT_Y$ may be written as the weighted average of the $ITT_Y$ effects across the four sub-populations of compliers, defiers, always-takers and never-takers:
    \begin{equation} \small
     ITT_Y = \pi_C ITT_{Y,C} + \pi_D ITT_{Y,D} + \pi_{NT} ITT_{Y,NT} + \pi_{AT} ITT_{Y,AT},
    \end{equation}
where $ITT_{Y,G}$ ($G=C, D, NT, AT)$ is the effect of the treatment assignment on units
    of type $G$ and $\pi_G$ is the proportion of units of type $G$.
    
   We can then proceed by adding step by step the four assumptions. 
    The first assumption that we impose is the exclusion restriction. If it holds, then we get
 \begin{equation} \small
 ITT_{Y,AT}=ITT_{Y,NT}=0,
 \end{equation}  
  since, for both always-takers and never-takers, one has
    \begin{equation} \small
    	Y_i(1)-Y_i(0)=0.
    \end{equation}
If for an individual the assignment has no effect on the treatment
    received, then it has also no effect on the outcome. This is a
    substantial assumption, and is not implied by the design. It is generally
    stated as the assignment not affecting the outcome other than through
    the treatment received, as we saw above.
 Such an assumption can be used
    to attribute the effect of assignment to the treatment received as follows, taking into account
		only compliers and defiers: 
   \begin{equation} \small
   ITT_Y = \pi_C ITT_{Y,C} +  \pi_D ITT_{Y,D}.
   \end{equation}
Under monotonicity, we rule out the existence of defiers: $\pi_{D}=0$.
    If we add the unconfoundedness assumption, we can estimate the distribution of compliance types as follows:
    \begin{enumerate}[a.]	
    	\item $\pi_{AT} = P(W_i(0) = W_i(1) = 1) =\mathbb{E}[W_i|Z_i = 0]=\mathbb{E}[W_i(0)]$, estimated as $\hat{\pi}_{AT}=\frac{1}{N_0}\sum_{i=1}^N (1-Z_i)W_i$;
    	
    	\item $
    	\pi_{NT} =P(W_i(0) = W_i(1) = 0) = 1 - \mathbb{E}[W_i|Z_i = 1]=1-\mathbb{E}[W_i(1)]
    	$, estimated as  $\hat{\pi}_{NT}=\frac{1}{N_1}\sum_{i=1}^N Z_i
    	(1-W_i)$;
    	
    	\item $
    	\pi_C = P(W_i(0) = 0, W_i(1) = 1) = \mathbb{E}[W_i|Z_i = 1]-\mathbb{E}[W_i|Z_i = 0]
    	$, estimated as $\hat{\pi}_{C}=\frac{1}{N_1}\sum_{i=1}^N Z_i W_i-
    	\frac{1}{N_0}\sum_{i=1}^N (1-Z_i)W_i$,
\end{enumerate}    	
where $N_1$ is the number of units assigned to the treatment and $N_0$ is the number of units assigned to the control.
Once one has estimated the distribution of compliers, when one adds also the ``existence of compliers'' assumption, one finally gets:	
    \begin{equation} \small
ITT_Y = \pi_C ITT_{Y,C}.
    \end{equation}
From this formula, as being $\pi_C \neq 0$,  it comes out that $ITT_{Y,C}$, the so-called Complier Average Causal Effect (CACE), is [\ref{imbens1997}]:
    \begin{equation} \label{ITT_C} \small
    ITT_{Y,C} = ITT_Y/\pi_C=\frac{\mathbb{E}[Y_i|Z_i = 1]-\mathbb{E}[Y_i|Z_i = 0]}{\mathbb{E}[W_i|Z_i =
    	1]-\mathbb{E}[W_i|Z_i = 0]}.
    \end{equation}
    In general, the global $ITT_Y$ may be viewed as a lower bound on 
    the treatment effect on the compliers: with the assumptions $\pi_D =
    0$, $\pi_C > 0$, and that both $ITT_{Y,NT}$ and $ITT_{Y,AT}$ are strictly less than
    $ITT_{Y,C}$, one gets $ITT_Y < ITT_{Y,C}$.
    The complier average treatment effect, $ITT_{Y,C}$, is a local effect, since it makes reference just to the population of compliers, hence it can also be referred as a Local Average Treatment Effect (LATE). It can be estimated as the coefficient associated with the instrumental variable regression  [\ref{imbens1994}] as we will see in detail in Subsection III.B.
	Invoking unconfoundedness, exclusion restriction and
    monotonicity, one can also infer the outcome distribution for
    compliers, $\mathbb{E}[Y_i(0)|G_i=C]$, and $\mathbb{E}[Y_i(1)|G_i=C]$. Under the same assumptions, one can estimate the entire marginal distribution of $Y_i(0)$ and $Y_i(1)$ for compliers.

     \section{Causal Tree with Instrumental Variable} \label{unconfounded}

    \textbf{A. Causal Tree with Randomized Instrumental Variable.} \label{CT-IVrandomized}
	In the following, we  extend the CT algorithm to the case of an irregular assignment mechanism where the assignment-to-the-treatment variable is itself randomized, but its receipt is not.
		If we assume the instrumental variable to be randomized, we can draw causal inference from a Causal Tree by making some changes in the structure of the tree. The first difference is that we need to rework the outcome variable, substituting in (\ref{eq:10}) the indicator variable $W_i$ with the instrumental variable $Z_i$, as follows:
    \begin{equation} \small
    Y_i^{*,IV}=Y_i^{obs} \cdot {{Z_i- e(X_i)}\over{(1-e(X_i)) \cdot e(X_i)}},
    \end{equation}
	where the propensity score is now reworked as $e(x) = P(Z_i=1|X_i=x)$.
    In this case, when the assignment mechanism corresponds to a classical randomized experiment, 
		the propensity score is a constant (i.e., $e(x)=p$ for all $x$), and the transformation above simplifies to:
    \begin{equation} \small
    Y_i^{*,IV}= {{Y_i(1) \cdot Z_i} \over p}- {{Y_i(0) \cdot (1 - Z_i) } \over {(1-p)}}.
    \end{equation}
    Likewise in (\ref{cace_leaf}), one can also use a weighted version of the transformation of the outcome to provide an estimate $\reallywidehat{ITT}_Y(X_i)$ of the intention to treat $ITT_Y$, for $X_i$ belonging to a generic leaf $\mathbb{X}_j$, as follows:
    \begin{equation} \label{IV} \footnotesize
    \reallywidehat{ITT}_Y(X_i)={{\sum_{l: X_l^{tr} \in \mathbb{X}_j}  Y_l^{obs,tr} \cdot {Z_l^{tr} \over \hat{p}}} \over {\sum_{l: X_l^{tr} \in \mathbb{X}_j}{Z_l^{tr} \over \hat{p}}}}-
    {{\sum_{l: X_l^{tr} \in \mathbb{X}_j}  Y_l^{obs,tr} \cdot {(1-Z_l^{tr}) \over (1-\hat{p})}} \over {\sum_{l: X_l^{tr} \in \mathbb{X}_j}{(1-Z_l^{tr}) \over (1-\hat{p})}}}, 
    \end{equation}
    where $\hat{p}$ is the estimated value of $p$. Again, following 
		[\ref{hirano2003}], (\ref{IV}) is an unbiased and efficient estimator of (\ref{super_IV}) within every leaf. 
    
    We also need to rework the in-sample and out-of-sample goodness-of-fit measures:
    \begin{enumerate}[a.]
    	\item In-sample goodness of fit:
    	\begin{equation}\label{IS_IV} \small
    	Q^{is, IV}=-{1\over N^{tr}}\sum_{i=1}^{N^{tr}} \reallywidehat{ITT}_Y^2(X_i^{tr});
    	\end{equation}
    	\item Out-of-sample goodness of fit:
    	\begin{equation} \small
    	Q^{oos, IV}=-MSE=-{1\over N^{va}} \sum_{i=1}^{N^{va}}(Y_i^{va,*}-\reallywidehat{ITT}_Y(X_i^{va}))^2.
    	\end{equation}
    \end{enumerate}
    
		For the sake of clarity, here (and in the following subsection), to fit the instrumental variable framework, we have reworked the out-of-sample goodness-of-fit 
		based on TOT (see (\ref{catebased})). This rework could easily be adapted to other out-of-sample goodness-of-fit measures. 
		
		The last part of our algorithm based on the instrumental variable focuses on the estimation of the complier average treatment effects. As we highlighted before, by using the instrumental variable $Z_i$, we are substantially assuming four different types in our population: compliers, always-takers, never-takers, and defiers. As before, our interest lies on the effect on the compliers.
  Within every leaf, the complier average causal effect is:
    \begin{equation} \footnotesize
    {\tau}^{cace}_{\mathbb{X}_j}=\frac{\mathbb{E}[Y_i|Z_i=1, X_i \in \mathbb{X}_j] - \mathbb{E}[Y_i|Z_i=0, X_i \in \mathbb{X}_j]}{P(G_i=C|X_i \in \mathbb{X}_j)}={ITT_{Y,\mathbb{X}_j} \over \pi_{C,\mathbb{X}_j}}.
    \end{equation}
 This formula is analogous to (\ref{ITT_C}), and can be estimated in every leaf assuming the existence of compliers.  Then, ${\tau}^{cace}_{\mathbb{X}_j}$ can be estimated as:
    \begin{equation} \label{cace_leaf1} \small
    \hat{\tau}^{cace}_{\mathbb{X}_j}={\reallywidehat{ITT}_{Y,\mathbb{X}_j} \over \hat{  \pi}_{C,\mathbb{X}_j}},
    \end{equation}
where ${ITT}_{Y,\mathbb{X}_j}$ is estimated following (\ref{IV}), and ${  \pi}_{C,\mathbb{X}_j}$ can be estimated as: 
\begin{equation} \label{pi_c} \small
\hat{\pi}_{C,\mathbb{X}_j}=\frac{1}{N_{1, \mathbb{X}_j}}\sum_{i=1}^{N_{\mathbb{X}_j}} Z_i W_i-\frac{1}{N_{0, \mathbb{X}_j}}\sum_{i=1}^{N_{\mathbb{X}_j}} (1-Z_i)W_i,
\end{equation}
where $N_{1, \mathbb{X}_j}$ and $N_{0, \mathbb{X}_j}$ are the numbers of units assigned respectively to the treated and control group within a certain leaf $\mathbb{X}_j$, and $N_{\mathbb{X}_j}$ is the number of units within the leaf.
\vspace{0.3cm}

    \textbf{B. Causal Tree with Unconfounded Instrumental Variable.} \label{CT-IVunconfounded} Now, we extend the analysis above to the case of an irregular assignment mechanism, where both the assignment and receipt of the treatment are not randomized, but the assignment can be assumed to be unconfounded when conditioning on important covariates.
		When the instrumental variable is not randomized a priori, the property of unconfoundedness of the instrument does not necessarily hold.
    If we think of $Z_i$ as our assignment mechanism, then the unconfoundedness of the instrument holds when:
    \begin{equation} \label{eq:40} \small
    Z_i \independent ({Y}_i(0), {Y}_i(1)) | {X}_i.
    \end{equation}
     Due to the propensity score properties, 
		this assumption holds even conditioning on the propensity score:
       \begin{equation} \label{eq:41} \small
       Z_i \independent ({Y}_i(0), {Y}_i(1)) | e({X_i}).
       \end{equation} 
   
    When the assumption (\ref{eq:41}) holds, one can rework the transformed outcome variable in a similar way as in the previous subsection, obtaining \small
    \begin{align} 
    Y_i^*&=Y_i(1) \cdot {Z_i \over e(X_i)} - Y_i(0) \cdot {{(1-Z_i)} \over {(1-e(X_i))}}.
    \end{align} \normalsize
   Assuming that the exclusion restriction and the monotonicity assumptions
    hold, 
		it is possible to provide an estimate $\reallywidehat{ITT}_Y(X_i)$ of the intention to treat $ITT_Y$ for $X_i$ belonging to a generic leaf $\mathbb{X}_j$, as follows:
    \begin{eqnarray} \label{IV_propensity} \footnotesize
    \reallywidehat{ITT}_Y(X_i)&=&{{\sum_{l: X_l^{tr} \in \mathbb{X}_j}  Y_l^{obs,tr} \cdot {Z_l^{tr} \over \hat{e}(X_l^{tr})}} \over {\sum_{l: X_l^{tr} \in \mathbb{X}_j}{Z_l^{tr} \over \hat{e}(X_l^{tr})}}} \nonumber \\
		&& -
    {{\sum_{l: X_l^{tr} \in \mathbb{X}_j}  Y_l^{obs,tr} \cdot {(1-Z_l^{tr}) \over (1-\hat{e}(X_l^{tr}))}} \over {\sum_{l: X_l^{tr} \in \mathbb{X}_j}{(1-Z_l^{tr}) \over (1-\hat{e}(X_l^{tr}))}}},
    \end{eqnarray}
    where $\hat{e}(X_l^{tr})$ is the estimated value of $e(X_l^{tr})$. The difference between (\ref{IV}) and (\ref{IV_propensity}) is that, given the complete randomization of the instrument, in (\ref{IV}) the probability $\hat{e}(X_l^{tr})$ was fixed to $\hat{p}$ for any given unit, while in (\ref{IV_propensity}) the assignment-to-the-treatment probability is modelled by the estimated propensity score $\hat{e}(X_l^{tr})$. Finally, the complier average treatment effect in each leaf is still estimated using (\ref{cace_leaf1}), replacing (\ref{IV}) with (\ref{IV_propensity}) to determine the estimate $\reallywidehat{ITT}_{Y,\mathbb{X}_j}$.

    \subsubsection{Overall CACE}
\par Starting from all the leaves, one can reconstruct the overall effect over all of them as a weighted average of the estimates $\hat{\tau}^{cace}_{\mathbb{X}_j}$ over every leaf $\mathbb{X}_j$.
One can represent this weighted average as
\begin{equation}\label{eq:tau_cace_overall} \small
\hat{\tau}^{cace}_{overall}= \sum_{j=1}^{N^{l}} \frac{\hat{\tau}^{cace}_{\mathbb{X}_j}  \cdot N^{co}_{\mathbb{X}_j}}{N^{co}},
\end{equation}
where $N^{l}$ represents the number of leaves, $N^{co}_{\mathbb{X}_j}$ the number of compliers for every leaf $\mathbb{X}_j$, and $N^{co}$ the overall number of compliers in all the leaves.
One can also compute the proportion of compliers in every leaf $\mathbb{X}_j$ simply as:
\begin{equation} \small
\pi^{co}_{\mathbb{X}_j}= \frac{N^{co}_{\mathbb{X}_j}}{N^{co}}.
\end{equation}

    \subsubsection{Estimating CACE in Every Leaf with Two Stage Least Squares Regressions} \label{2sls}
    A suitable possibility to estimate the treatment effect in every leaf is to use, within every leaf $\mathbb{X}_j$
    of the tree $\mathbb{T}$, the Two Stage Least Squares (henceforth, TSLS) method for the estimation of the effect on the complier population, as it is presented in [\ref{imbens1997}]. If one assumes the receipt of the treatment variable $W_i$ and the instrumental variable $Z_i$ to be binary variables, our problem can be expressed in terms of 2 simultaneous regressions:
		
    \small
\begin{align} \label{IV_reg}
Y_i^{obs}&=\alpha + \gamma \cdot W_i + \epsilon_i, \\ \label{IV_reg1}
W_i&=\pi_0 + \pi_1 \cdot Z_i + \eta_i.
\end{align} \normalsize

In the econometric terminology, the explanatory variable $W_i$ is $endogenous$, while the IV variable $Z_i$ is $exogenous$.  
 \par The logic of IV regression is that one can estimate the above two reduced form regressions in the case of a single instrument by least squares. In particular, one can estimate  $\tau^{CACE}$ through TSLS, as the following ratio [\ref{ir2015}, \ref{imbens1997}]:
    \begin{equation} \label{eq:52} \small
    \hat{\tau}^{CACE}=\hat{\tau}^{IV}=\frac{\hat{\gamma}}{\hat{\pi}_1}=\frac{\reallywidehat{ITT}_Y}{\hat{  \pi}_C},
    \end{equation}
    where $\hat{\tau}^{IV}$ is an unbiased estimator of the average causal effect on the population of compliers.
    If one runs a TSLS regression within every leaf $\mathbb{X}_j$ of the tree $\mathbb{T}$, then one is able to obtain an estimate $\hat{\tau}^{cace}_{\mathbb{X}_j}$ for every such leaf.
    
    A possible extension of (\ref{IV_reg1}) would be to include in the first stage regression all the possible confounding variables available in the dataset (in this case, $\pi_2 \cdot X_i$ denotes a scalar product):
    \begin{equation} \small
    W_i=\pi_0 + \pi_1 \cdot Z_i + \pi_2 \cdot X_i + \eta_i.
    \end{equation}
     The idea is that, if the instrument is unconfounded only conditional on confounding variables, then one could include these covariates in the estimation of the treatment effect on the complier population within each leaf. 

    \par In every leaf, using the TSLS method, we can also obtain an estimate of the variance of our $\hat{\tau}^{cace}_{\mathbb{X}_j}$ estimator, which corresponds to the Neyman estimated variance $\hat{\mathbb{V}}^{Neyman}_{\mathbb{T},\mathbb{X}_j}$ for the leaf $\mathbb{X}_j$ of the tree $\mathbb{T}$ (see (\ref{Neyman})). 
    \vspace{0.3cm}

\textbf{C. The CT-IV Algorithm.} \label{algorithm} Our proposed CT-IV algorithm is summarized as follows.
  
 \line(1,0){240}
 
 \textbf {Causal Tree with Instrumental Variable (CT-IV)} \\
 \footnotesize 
 
 Inputs: $N$ units $i$ $(X_i, Z_i, W_i, Y_i^{obs})$, where $X_i$ is the feature vector, $Z_i$ is treatment assignment (instrumental variable), $W_i$ is the treatment receipt, and $Y_i^{obs}$ is the observed response.

Outputs: 1) a Causal Tree (determined by the use of the instrumental variable), and 2) estimates of the Complier Average Causal Effects on its leaves.

  \begin{enumerate}  
  	\item First Step of the Algorithm (Building the Tree)
  	\begin{itemize}
  		\item Draw a random subsample from $\Omega$ without replacement and divide it into two disjoint sets: a training set ($\Omega^{tr}$) and a validation set ($\Omega^{va}$) of size $\sum_{k}{\Omega_k}=\Omega$ with $k \in \{tr, va\}$.
  		\item Grow a Causal Tree, following the next procedure to take into account the presence of the instrumental variable $Z_i$:
  		\renewcommand{\theenumii}{\roman{enumii}}
  		\begin{enumerate}
  			\item estimate the propensity $e(x)=P(Z_i=1 | X_i = x)$ of getting assigned to the treatment;
  			\item drop units with an estimated propensity
  			score below 0.1 or above 0.9 (in order not to weight too much units with extreme values of the estimated propensity score); 
  			\item grow a tree by maximizing the following in-sample goodness-of-fit criterion, for several values of $\alpha > 0$:	\begin{equation*}
  			Q^{crit, IV}=-{1\over N^{tr}}\sum_{i=1}^{N^{tr}} \reallywidehat{ITT}_Y^2(X_i^{tr}) - \alpha\cdot \kappa,
  			\end{equation*}
			where $ \reallywidehat{ITT}_Y(X_i^{tr})$ is estimated on the training sample as in (\ref{IV}) in the case of randomization of the instrument or as in (\ref{IV_propensity}) if the instrument is not randomized, $\alpha$ is the penalty parameter, and $\kappa$ is the number of leaves, which measures the complexity of the model;
  			\item cross-validate the tree, using the following out-of-sample goodness of fit:
  		\begin{equation*}
  			Q^{oos, IV}=-MSE=-{1\over N^{va}} \sum_{i=1}^{N^{va}}(Y_i^{va,*} - \reallywidehat{ITT}_Y(X_i^{va}))^2
  			\end{equation*}
  			where $ \reallywidehat{ITT}_Y(X_i^{va})$ is estimated on the validation sample as in (\ref{IV}) in the case of randomization of the instrument or as in (\ref{IV_propensity}) if the instrument is not randomized.
  			\end{enumerate}
  			\end{itemize}
  				\item Second Step of the Algorithm (Estimating the Complier Average Causal Effects)
  			\begin{itemize}
  				\item The complier average causal effect within a leaf $\mathbb{X}_j$ can be estimated on the entire sample $\Omega$ in two alternative ways:
  				\begin{enumerate}[(a)]
  					\item if $Z_i$ is randomized (Subsection III.A) then one can directly estimate the complier average causal effect within every leaf as:
  					\begin{equation*}
  					\hat{\tau}^{cace}_{\mathbb{X}_j}={\reallywidehat{ITT}_{Y,\mathbb{X}_j} \over \hat{  \pi}_{C,\mathbb{X}_j}}
  					\end{equation*}
  					where $ \reallywidehat{ITT}_{Y, \mathbb{X}_j}$ is estimated as in (\ref{IV}) and $\hat{  \pi}_{C,\mathbb{X}_j}$ is estimated following (\ref{pi_c});
  					\item if $Z_i$ is not randomized but can be assumed to be unconfounded (Subsection III.B) then run a TSLS conditioning on the confounding covariates in the first stage regression.
  				\end{enumerate}
  			\end{itemize}	
  			

\end{enumerate}	
\line(1,0){255}
\vspace{0.1cm}
\normalsize

 \section{Comparison of the CT and CT-IV Algorithms on Synthetic Data} \label{synth}
 In this section, we conduct simulations on synthetic data, to compare the performance of the proposed CT-IV algorithm with that of CT. As goodness-of-fit measure, we use the opposite of the Mean Squared Error of prediction (MSE) on the test set, and to assess the relative performance of the two algorithms, we consider the following relative gap measure based on such MSE [\ref{wanglihopp2017}]: 

 \begin{equation} \small
 Relative\,Gap = \frac{MSE_{CT} - MSE_{CT\textit{-}IV}}{MSE_{CT}} \times 100.
 \end{equation}
 Moreover, we run some robustness checks. 
In this section, our focus will be also on what happens in presence of a weak instrument, namely when the instrument $Z_i$ is weakly correlated with the treatment variable, and when the instrument directly affects the outcome.
	While the presence of weak instruments is directly testable (typically, with an F-test on the first stage regression), what is not testable and could be potentially harmful is a violation of the exclusion restriction at the leaf level. Alternative algorithms, such as the one 
	in [\ref{wanglihopp2017}], take into account the exclusion restriction just at a general level while, in this paper, we take into account that assumption at the leaf level. In a non-synthetic scenario, this assumption is not directly testable, but our algorithm seems to be more ``transparent''  than other algorithms by taking into account possible violations of this assumption. \vspace{0.3cm}
 
 \textbf{A. Synthetic Data Construction.}
To compare our CT-IV algorithm with the CT one, we first consider some scenarios where the assignment mechanism is irregular. As we saw in Subsection II.C, this means that the assignment to the treatment is randomized, but the receipt of the treatment is not.
The general model that we use for our data simulation is built by considering the following variation of the typical IV setting reported in (\ref{IV_reg}) and (\ref{IV_reg1}). The major differences are that we introduce in the main equation (\ref{IV_reg3}) a nuisance term $\eta_i$ and an interaction term between regressors and the treatment indicator, in order to \textit{heterogenise} the treatment effects. The nuisance term $\eta_i$ can be thought as a not-observable feature that affects both the treatment assignment and the outcome. The general setting looks as follows:
\vspace{-0.4cm}

{\small
  \begin{align} \label{IV_reg3} 
  Y_i^{obs}&= 1 + f(X_i^{out}) +  W_i +  W_i \cdot g(X_i^{tre}) + \eta_i + \epsilon_i, \\ \label{IV_reg2}
  W_i&= Z_i +f(X_i^{out}) + \eta_i,
  \end{align}
	}
	\normalsize
\noindent where $X_i = (x_{i1},..., x_{iK})$ is a $K$-dimensional vector of covariates, $X^{out}_i$ highlights those covariates that have an effect on the outcome, and $X_i^{tre}$  (with $\{X_i^{tre}\}\subseteq \{X^{out}_i\}$) those covariates that affect the treatment effect.
  We consider various functional forms for $f$ and $g$ and for the error distribution in the main equation (\ref{IV_reg3}), as well as for $f$ in the first stage equation (\ref{IV_reg2}). 
	The 
	designs 
	investigated (with $K=1$ for design 1, and $K=10$ for the other cases) are reported in Table \ref{sim}.

    \begin{table}
		\scriptsize
    	\centering
    	\caption{Simulation models} \label{sim}
  		\begin{tabular}{ p{0.3cm}p{6.1cm}p{0.6cm}  } 
  			\toprule
  			\textbf{\textbf{Design}}  & \centering \textbf{Form of the Model} & \textbf{Error}  \\  
  			\midrule
  			1 & $Y_i^{obs}= 1 + X_{i1} +  W_i +  W_i \cdot X_{i1} + \eta_i + \epsilon_i$ & $ \mathcal{N}(0,1)$   \\
  			2 & $Y_i^{obs}= 1 + \sum_{k=1}^{10} X_{ik} +  W_i +  W_i \cdot \sum_{k=9}^{10}X_{ik} + \eta_i + \epsilon_i$ & $ \mathcal{N}(0,1)$  \\
  			3 & $Y_i^{obs}= 1 + \sum_{k=1}^{10} X_{ik} +  W_i +  W_i \cdot \sum_{k=9}^{10}X_{ik} + \eta_i + \epsilon_i$ & $ Exp(10)$  \\  
  			4 & $Y_i^{obs}= 1 + \sum_{k=1}^{10} X_{ik} +  W_i +  W_i \cdot \sum_{k=9}^{10}X_{ik} + \eta_i + \epsilon_i$ & $\mathcal{U}(0,1)$   \\
  			5 & $Y_i^{obs}= 1 + \sum_{k=1}^{10} X_{ik} +  W_i +  W_i \cdot \prod_{k=9}^{10}X_{ik} + \eta_i + \epsilon_i$ & $ \mathcal{N}(0,1)$  \\
  						\bottomrule
  			\multicolumn{3}{l}{} \\
  		\end{tabular} \vspace{-0.6cm}
        \end{table}
We train all the five models using incrementally bigger samples, with cardinality ranging from 500 to 50000 (i.e., 500, 1000, 5000, 50000). We implement a holdout cross-validation, assigning half of the observations to the training set (and validation set) and the other half to the test set. We let $X_{ik} \sim\mathcal{N}(0,0.1)$ (considering independent features), $W_i \sim Bern(0.5)$, $Z_i \sim Bern(0.5)$, and the nuisance parameter $\eta_i$ be a white noise. Moreover, we set the correlations between $W_i$ and $Z_i$ and $W_i$ and $\eta_i$ to be respectively $Cor(W_i, Z_i) \simeq 0.65$ and $Cor(W_i, \eta_i) \simeq 0.50$. To make the trees comparable, we set the maximal depth of the tree to be 2, and the minimal leaf size to be one tenth of the sample size. \vspace{0.3cm}

 \textbf{B. Simulation Results.}
The results of the simulations are evaluated in Table \ref{sim_res}, in terms of the mean squared error of prediction on the test set. 	As one can see from the relative gaps reported in Table \ref{sim_res}, the IV-CT algorithm outperforms the CT one in all the different designs. Comparing the various models by column, one observes that with respect to the baseline case (design 1), the relative gap between the IV-CT and CT algorithms widens as we add covariates (design 2), change the errors distribution (designs 3 and 4), or change the functional form (design 5).
Moreover, it is important to notice that, as the sample size increases, the relative gap widens as well. From the values of the MSE it seems that, while the CT performance is quite stable, CT-IV performance increases as the sample size grows larger. This is especially true in designs 1 and 5.

\begin{table}
\scriptsize
		\centering
        \caption{Simulation results} 
		\label{sim_res}
		\begin{tabular}{c c c c c c }
			\toprule
			\multirow{2}{*}{\textbf{Design}} & \multirow{2}{*}{\textbf{Approach}} & \multicolumn{4}{c}{\textbf{Sample Size}} \\
			\cline{3-6}
			& & \multicolumn{1}{c}{500} & \multicolumn{1}{c}{1,000} & \multicolumn{1}{c}{5,000} & \multicolumn{1}{c}{50,000}  \\
			\midrule
			\multirow{3}{*}{1} & MSE (CT-IV) & 0.369 & 0.038 & 0.067 & 0.066  \\
			& MSE (Causal Tree) & 0.857 & 0.727 & 1.073 & 0.973   \\
			& Relative Gap & 57\% & 94\%  & 94\% & 93\%   \\
			\midrule
				\multirow{3}{*}{2} & MSE (CT-IV) & 0.239 & 0.058 & 0.058  & 0.052 \\
			& MSE (Causal Tree) & 0.778 & 0.787 & 1.028 & 0.994    \\
			& Relative Gap & 69\% &  93\% & 94\% &  95\%  \\
			\midrule
				\multirow{3}{*}{3} & MSE (CT-IV) & 0.058  & 0.041 & 0.037 & 0.062 \\
			& MSE (Causal Tree) & 0.872 & 1.186 & 1.044 & 1.004   \\
			& Relative Gap & 93\% & 97\%  &  96\%&  94\%    \\
			\midrule
				\multirow{3}{*}{4} & MSE (CT-IV) & 0.072 &  0.053 & 0.051 & 0.050  \\
			& MSE (Causal Tree) & 0.851  & 1.052 & 1.098 & 1.004  \\
			& Relative Gap & 92\% & 95\% & 95\% & 95\% \\
			\midrule
				\multirow{3}{*}{5} & MSE (CT-IV) & 0.122 & 0.030 & 0.070 & 0.058   \\
			& MSE (Causal Tree) & 0.893 & 0.866 & 0.720 & 1.014    \\
			& Relative Gap & 86\% & 96\% & 90\% & 94\%   \\
			\bottomrule
		\end{tabular} \vspace{-0.4cm}
	\end{table} \vspace{0.3cm}
	
 \textbf{C. Robustness Checks.} 
Once we have checked that the CT-IV algorithm outperforms the CT algorithm on synthetic data, it is worth asking what happens when some of the assumptions on which the consistency of the CT-IV is built are partially violated. 
The main problem that can arise when applying the CT-IV method is a well-known issue in the econometric literature, known as the \textit{weak instrument} problem [\ref{stock2002}]. 
This problem, in our framework, deals with the fact that the number of compliers within every leaf can be particularly small. In an econometric framework, the goal that one would like to achieve is to ensure that  $Cor(W_i, Z_i)$ is bounded away from zero. 
In the following, we test what happens when the instrument is weak on the overall population, and what happens when the exclusion restriction is violated in a specific sub-population. 
We test these violations on the second model design in Table \ref{sim}. In particular, we assume 2 different scenarios. In the first scenario, we let the instrument be weak on the overall population, by setting $Cor(W_i, Z_i) \simeq 0.5 $.
 In the second scenario, we impose a partial violation of the exclusion restriction, by letting the instrumental variable $Z_i$ directly affect the outcome $Y_i$ when the feature $X_{i10}$ satisfies the condition $ X_{i10} \geq 0$. 
\begin{table}
\scriptsize
	\centering
    \caption{Robustness checks for violations of IV assumptions} 
	\label{rc_res}
	\begin{tabular}{c c c c c c}
		\toprule
		\multirow{2}{*}{\textbf{Scenario}} & \multirow{2}{*}{\textbf{Approach}} & \multicolumn{4}{c}{\textbf{Sample Size}} \\
		\cline{3-6}
		& & \multicolumn{1}{c}{500} & \multicolumn{1}{c}{1,000} & \multicolumn{1}{c}{5,000} & \multicolumn{1}{c}{50,000}  \\
		\midrule
		\multirow{3}{*}{1} & MSE (CT-IV) & 0.439 & 0.157 & 0.120 & 0.194  \\
		& MSE (Causal Tree) & 0.881 & 0.898 & 1.270  & 1.252 \\
		& Relative Gap & 50\% & 82\%  & 90\% & 85\% \\
		\midrule
		\multirow{3}{*}{2} & MSE (CT-IV) & 0.198 & 0.040 & 0.118 & 0.143 \\
		& MSE (Causal Tree) & 0.244 & 0.329 & 0.452 & 0.311  \\
		& Relative Gap & 19\% &  87\% & 74\% & 54\% \\
		\bottomrule
	\end{tabular} \vspace{-0.4cm}
\end{table}

In this case, the results from the simulations, reported in Table \ref{rc_res}, show that the IV-CT algorithm outperforms the CT even in the presence of weak instruments. It is important to notice that, within every leaf, the weak-instrument test leads to the rejection of the null hypothesis of weak instrument: our algorithm is able to identify those leaves where there is no weak-instrument problem. 
Moreover, our algorithm is robust even when the exclusion restriction is partially violated (second scenario). In this case, while the CT algorithm shows a better performance compared with the other scenario, by partially reducing the relative gap, the CT-IV still performs better in terms of the mean squared error of prediction. Since the estimation of the causal effects is performed in a second stage with respect to the building of the tree, our algorithm seems to handle in a good way possible problems due to the violation of exclusion restriction within every leaf. This could not hold true if the exclusion restriction is taken into account just at a general level, as in [\ref{wanglihopp2017}]. 

\section{Case Study} \label{assess}

\textbf{A. Programs for the Development of Crafts in Tuscany (Italy).} During the years 2003-2005, the Tuscan Regional Administration (Italy) introduced the ``Programs for the Development of Crafts'' (henceforth, PDC). These programs were aimed at Tuscan small-sized handicraft firms, with the goal of promoting innovation and regional development [\ref{arpino2016}, \ref{mealli2017}].  The firms could access PDC by a voluntary application and eligibility criteria. The objective of PDC was to ease access to credit for small-sized firms to boost investments, sales and employment levels.
 The PCD call guaranteed soft-loans to the firms that were considered eligible for the grant. The eligibility was evaluated on the basis of an investment project.
 The minimal admissible investment cost was 25\,000 Euros, and the grant covered 60\% of the financed investment [\ref{mattei2007}].
Among firms participating in the PDC, the large majority of the projects were funded, and the percentage of insolvencies was lower than 3\%.
Data are available for firms that received the PDC, firms that applied for the founding but were not eligible, and firms that did not apply for the PDC. For our analysis, we use an integrated dataset including information collected by the ``Artigian Credito Toscano'' and information coming from the archives of the Chamber of Commerce ($2001-2004$).
The data are available for $266$ assisted firms (participating in 2003/05 PDC) and $721$ non-assisted firms.
The firms in the dataset are operating in 4 economic sectors that comprise the majority of the Tuscan artisan firms:
 construction; manufacturing activities; wholesale and retail trade; real estate business, rental services, computer, research, business services.
The covariates are time-varying covariates, such as sales and employees, and time-invarying covariates, such as location of the firm, year of start-up, legal status, and main distribution channel.
The location of each firm is recorded at the provincial level. 
A central variable for our analysis is the amount of firm's sales in 2002 (pre-treatment year). We created 6 different sales' groups (up to 50,000; 50,000 to 100,000; 100,000 to 250,000; 250,000 to 500,000; 500,000 to 1,000,000; greater than 1,000,000).
The dependent variable is a categorical variable that takes the values $1$, $-1$, $0$, respectively, if the number of employees in the firm in the year immediately subsequent to the treatment increased, decreased, or remained the same. 
 The covariate that catches the assignment to the treatment is a dummy variable, which is recorded as 1 if the firm received the financial aid during the two years 2003/2005, and 0 otherwise. However, the treatment variable is not randomized. To draw proper causal inference in this scenario (irregular assignment mechanism), one needs to use an instrumental variable. Luckily, we have data on firms' applications for the funding. 
This is a good instrument, since those firms that applied for the funding were very likely to get it, and it seems that the application itself should not have affected the outcome (exclusion restriction). Moreover, we know that the population of compliers exists (existence of compliers) and, since this is a case of one-sided non-compliance, defiers are ruled out by the design of the policy (monotonicity). 
However, the instrument is not randomized and, in order to draw proper causal inference in this scenario, one needs to build a propensity score for the instrument itself, in order for the unconfoundedness assumption to hold. This is a very central point and a main novelty of the approach proposed in this paper. 
Finally, we discard firms with missing values on relevant variables.
The selection procedure leads to a sample of 98 assisted firms and 662 non-assisted firms. \vspace{0.3cm}

\begin{figure}
	\centering
	\includegraphics[width=0.57\textwidth, center]{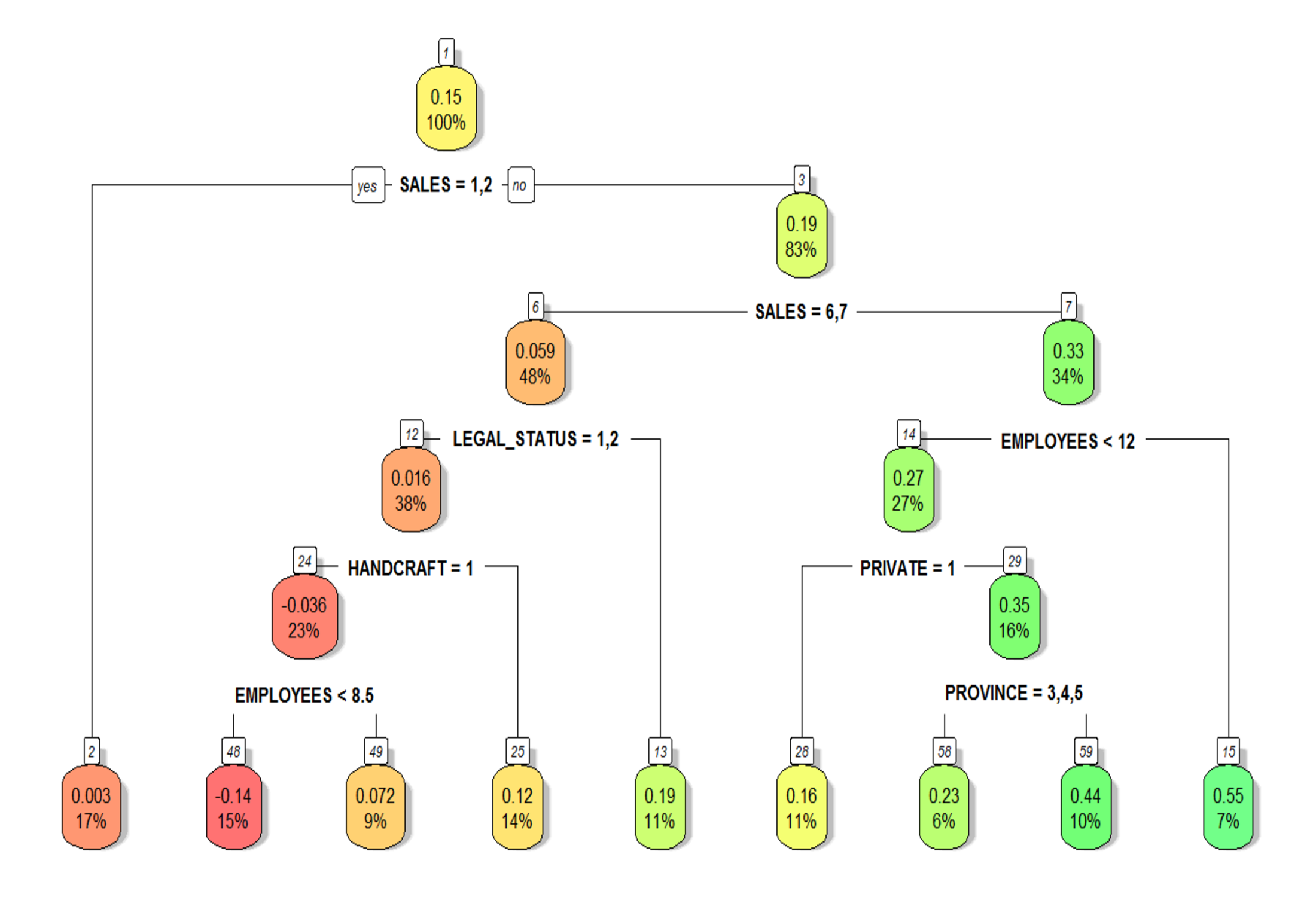}
	\caption{CT-IV built on the PDC data}
	\label{cartiv}
\end{figure}

 \textbf{B. Application of the CT-IV Algorithm to Assess Heterogeneous Causal Effects.} In Figure \ref{cartiv}, results are shown for the CT-IV algorithm applied on our data. Within every node, the estimated value of the intention to treat and the percentage of observations associated with each node are shown. The name of the variable used for splitting the tree is shown just below each node. The nodes are numbered according to their level in the tree (with the same numbering that one would use for a full binary tree with the same number of levels). The different effects are recorded with different colours: the pinker the node, the closer the effect to zero, while the greener the node, the stronger the effect. No weak-instrument problems occur in the construction of the tree. Table \ref{table:4} reports the estimated values of the ITT and of the CACE, for each leaf of the constructed tree. For the latter, we report also its standard error, estimated using the TSLS method within every leaf. As one can see, there is a large variation in the CACE among the different leaves. For instance, the observations in the \textit{node 2} (those firms with sales lower than 100,000 Euros) provide a null and non-significant effect according to a t-test. On the opposite side, those firms within the \textit{node 15} (firms with more than 11 employees, and with sales between 100,000 and 1,000,000 Euros) show a 0.61 causal effect in the chance of increasing the number of employees when they get the fundings. In our case study, the difference between $\reallywidehat{ITT}_{Y,{\mathbb{X}_j}}$ and $\hat{\tau}^{cace}_{\mathbb{X}_j}$ is quite small,
 because the percentage of compliers within every leaf is around 90\% (with the exception of \textit{node 58}, where the percentage of compliers is smaller than 80\%), and the overall percentage of compliers is exactly 90\%.
The overall CACE is positive (0.17) and significant. 

\begin{table}
 	\footnotesize \footnotesize
 	\centering
    \caption{Estimated Intention to Treat and Complier Average Causal Effect on the final nodes (leaves)}
 	\setlength{\tabcolsep}{2pt}
\renewcommand{\arraystretch}{1.5}
\begin{tabular}{rrrrrrrrrr}
 		\hline
 		Node $\#j$ &\#2& \#48& \#49& \#25& \#13& \#28& \#58& \#59& \#15 \\
 		\hline
 		$\reallywidehat{ITT}_{Y,{\mathbb{X}_j}}$ & 0.003 & -0.14 & 0.072 & 0.12 & 0.19 & 0.16 & 0.23 & 0.44 & 0.55 \\
 		$\hat{\tau}^{cace}_{\mathbb{X}_j}$ & 0.003 & -0.15 & 0.083 & 0.14 & 0.21 & 0.18 & 0.30 & 0.47 & 0.61 \\
 		S.E. $\hat{\tau}^{cace}_{\mathbb{X}_j}$ & 0.060 & 0.08 & 0.101 & 0.09 & 0.10 & 0.10 & 0.14 & 0.10 & 0.10 \\
 		\hline
 	\end{tabular} \vspace{-0.4cm}
	\label{table:4}
 \end{table}

 

 \section{Conclusions}
 The main aim of this paper is to strengthen the link between machine learning techniques and causal inference, as well as to provide, in this regard, an innovative approach.
From this point of view, the CT-IV algorithm developed in this paper has shown to fit in a good way our causal inference goals in the presence of an irregular assignment mechanism. The results obtained from the simulations show that, on one side, CT-IV provides a robust estimation of the overall causal effect on the population under study. On the other side, it outperforms the Causal Tree, providing a very good insight into the heterogeneity of the effects. As a possible extension, which will the subject of a future investigation, the CT-IV algorithm could be combined with the Honest Causal Tree framework [\ref{atheyimbens2016}], to improve the quality of its estimates.

Studying the heterogeneity of causal effects is growing in importance as the size of the dataset, and thus of the population under study, grows; indeed, as shown in the case study, taking into account heterogeneous effects on different subgroups of the study population can help optimizing public interventions and making them more cost-effective. Other possible applications are in fields such as management, health sciences, economics, sociology, and political science.

Concluding, CT-IV  has the following peculiar strengths: 
\begin{enumerate}
\item it can be applied directly, even when the instrument is not randomized (confounded instrumental variable);
\item it does not directly need a theoretical derivation of the consistency of its estimators because it is based on robust estimators (ITT \& TSLS estimators);
\item it provides robust causal effect estimators within every leaf, even when the exclusion restriction is partially violated;
\item it is also robust in settings where the instrument is weak. 
\end{enumerate}

\section*{Acknowledgment}
The authors declare no conflicts of interest.


\begin{thebibliography}{12} 
 	
 	
 	 \bibitem{itemreference4} Arpino, B., and Mattei, A. (2016). Assessing the Causal Effects of Financial Aids to Firms in Tuscany Allowing for Interference. {\em The Annals of Applied Statistics} 10(3), 1170-1194. \label{arpino2016}
    


 	
 	\bibitem{itemreference7} Athey, S., and Imbens, G.W. (2016). Recursive Partitioning for Heterogeneous Causal Effects. {\em Proceedings of the National Academy of Sciences} 113(27), 7353-7360. \label{atheyimbens2016}
 	
 	\bibitem{itemreference8} Athey, S., and Imbens, G.W. (2015). Machine Learning Methods for Estimating Heterogeneous Causal Effects. {\em Stat} 1050(5). \label{atheyimbens2015}
 

  	\bibitem{itemreference13} Breiman, L. (2001). Random Forests. {\em Machine Learning} 45(1), 5-32. \label{breiman2001}
 	
    \bibitem{itemreference14} Breiman, L., Olshen, J.H., Stone, C.J. (1984). {\em Classification and Regression Trees}. CRC Press. \label{breiman1984}
   
	
 	
 	\bibitem{itemreference27} Hirano, K., Imbens, G.W., and Ridder, G. (2003). Efficient Estimation of Average Treatment Effects using the Estimated Propensity Score. {\em Econometrica} 71(4), 1161-1189. \label{hirano2003}
 	
 	
 	
 	
 	\bibitem{itemreference32} Imbens, G.W., and Angrist, J.D. (1994). Identification and Estimation of Local Average Treatment Effects. {\em Econometrica} 62(2), 467-475. \label{imbens1994}
 	
    \bibitem{itemreference34} Imbens, G.W., Rubin, D.B. (2015). {\em Causal Inference for Statistics, Social, and Biomedical Sciences. An Introduction}. Cambridge University Press. \label{ir2015}
    
       \bibitem{itemreference33} Imbens, G.W., and Rubin, D.B., (1997). Estimating Outcome Distributions for Compliers in Instrumental Variables Models. {\em The Review of Economic Studies} 64(4), 555-574. \label{imbens1997}
       
     \bibitem{itemreference36} Mattei, A., and Mauro, V. (2007). {\em Valutazione di Politiche per le Imprese Artigiane}. Research Report. IRPET - Istituto Regionale Programmazione Economica della Toscana. \label{mattei2007}
     
     \bibitem{itemreference37} Mariani, M., and Mealli, F. (2018). The Effects of R\&D Subsidies to Small and Medium-Sized Enterprises. Evidence from a Regional Program. {\em Italian Economic Journal}, 4(2), 249-281. \label{mealli2017}
 
 	\bibitem{itemreference38} Neyman, J. (1934). On the Two Different Aspects of the Representative Method: The Method of Stratified Sampling and the Method of Purposive Selection. {\em Journal of the Royal Statistical Society} 97(4), 558-625. doi:10.2307/2342192 \label{neyman1934}

 	\bibitem{itemreference40} Pearl, J. (2009). {\em Causality}. Cambridge University Press. \label{pearl2009}
 	
 	\bibitem{itemreference42} Rosenbaum, P., and Rubin, D.B. (1983). Assessing the Sensitivity to an Unobserved Binary Covariate in an Observational Study with Binary Outcome. {\em Journal of the Royal Statistical Society, Ser. B} 45(2), 212-218.  \label{rosenbaum1983}
\vspace{-0.3cm}
    \bibitem{itemreference44} Rubin, D.B. (1974). Estimating Causal Effects of Treatments in Randomized and Nonrandomized Studies. {\em Journal of Educational Psychology} 66(5), 688-701. \label{rubin1974}
    
    \bibitem{itemreference} Rubin, D.B. (1980). Randomization analysis of experimental data: The Fisher randomization test comment. {\em Journal of the American Statistical Association} 75(371), 591-593. \label{rubin1980}

 	\bibitem{itemreference50} Stock, J.H., and Yogo, M. (2002). Testing for Weak instruments in Linear IV Regression. In Andrews, D.W.K., {\em Identification and Inference for Econometric Models}. New York: Cambridge University Press, 80-108. \label{stock2002}
 	
 	\bibitem{itemreference52} Wager, S., and Athey, S. (2017). Estimation and Inference of Heterogeneous Treatment Effects using Random Forests. {\em Journal of the American Statistical Association}, to appear. \label{wagerathey2017}
 	
 	\bibitem{itemreference53} Wang, G., Li, J. and Hopp, W.J., (2017). An Instrumental Variable Tree Approach for Detecting Heterogeneous Treatment Effects in Observational Studies. Technical Report: \url{https://papers.ssrn.com/sol3/papers.cfm?abstract_id=3045327}. \label{wanglihopp2017}

 	
 \end{thebibliography}
\end{document}